\documentclass{article} 
\usepackage{iclr2024_conference,times}

\usepackage{amsmath,amsfonts,bm}

\def\eqref#1{equation~\ref{#1}}

\def\1{\bm{1}}

\DeclareMathAlphabet{\mathsfit}{\encodingdefault}{\sfdefault}{m}{sl}
\SetMathAlphabet{\mathsfit}{bold}{\encodingdefault}{\sfdefault}{bx}{n}

\def\gL{{\mathcal{L}}}

\usepackage{hyperref}
\usepackage{url}
\pdfoutput=1

\usepackage{epsfig}
\usepackage[linesnumbered,boxed,ruled,commentsnumbered]{algorithm2e}
\usepackage{algorithmicx}
\usepackage{graphicx}
\usepackage{subfig}
\usepackage{makecell}
\usepackage{sidecap}
\usepackage{color}

\usepackage{booktabs}
\usepackage{multirow}
\usepackage{multicol}
\usepackage{makecell}
\usepackage{enumitem}
\usepackage{tikz}

\usepackage{times}
\usepackage{latexsym}

\usepackage[T1]{fontenc}
\usepackage[utf8]{inputenc}

\usepackage{microtype}
\usepackage{inconsolata}

\usepackage{bbding}

\newcommand{\bW}{\boldsymbol{W}}

\newcommand{\bX}{\boldsymbol{X}}

\newcommand{\bx}{\boldsymbol{x}}

\newcommand{\bh}{\boldsymbol{h}}

\newcommand{\bg}{\boldsymbol{g}}

\newcommand{\bb}{\boldsymbol{b}}

\newcommand{\bQ}{\boldsymbol{Q}}

\newcommand{\bV}{\boldsymbol{V}}

\newcommand{\bA}{\boldsymbol{A}}
\newcommand{\bB}{\boldsymbol{B}}

\newcommand{\bE}{\boldsymbol{E}}

\title{Delta-LoRA: Fine-Tuning High-Rank Parameters with the Delta of Low-Rank Matrices}

\author{Bojia Zi$^{1,2,3}$ ,
  Xianbiao Qi$^{3}$,
  Lingzhi Wang$^{1,2}$,
  Jianan Wang$^{3}$,
  Kam-Fai Wong$^{1,2}$\&
  Lei Zhang$^{3}$ \\
  $^{1}$The Chinese University of Hong Kong, Hong Kong, China \\
  $^{2}$MoE Key Laboratory of High Confidence Software Technologies, China \\
  $^{3}$International Digital Economy Academy (IDEA), Shenzhen, Guangdong, China \\
  \texttt{\{bjzi, lzwang, kfwong\}@se.cuhk.edu.hk} \\
  \texttt{\{qixianbiao, wangjianan, leizhang\}@idea.edu.cn}
  }
\begin{document}
\maketitle
\begin{abstract}
In this paper, we present \textbf{Delta-LoRA}, which is a novel parameter-efficient approach to fine-tune large language models (LLMs). In contrast to LoRA and other low-rank adaptation methods such as AdaLoRA, Delta-LoRA not only updates the low-rank matrices $\bA$ and $\bB$, but also propagate the learning to the pre-trained weights $\bW$ via updates utilizing the delta of the product of two low-rank matrices ($\bA^{(t+1)}\bB^{(t+1)} - \bA^{(t)}\bB^{(t)}$). Such a strategy effectively addresses the limitation that the incremental update of low-rank matrices is inadequate for learning representations capable for downstream tasks. Moreover, as the update of $\bW$ does not need to compute the gradients of $\bW$ and store their momentums, Delta-LoRA shares comparable memory requirements and computational costs with LoRA. Extensive experiments show that Delta-LoRA significantly outperforms existing low-rank adaptation methods. We further support these results with comprehensive analyses that underscore the effectiveness of Delta-LoRA.
\end{abstract}

\section{Introduction}
Large Language Models (LLMs) recently have attracted considerable attention due to their remarkable performance across a broad spectrum of downstream tasks. Diverging from conventional Transformers characterized by a scale of {millions of parameters, modern LLMs typically scale up to billions of parameters, endowing them with notable advantages such as emergent capabilities and robust generalization as detailed in \citep{gpt4_bubeck2023sparks}. 
Fine-tuning such highly capable LLMs on downstream tasks~\citep{t5_raffel2020exploring,bert_devlin-etal-2019-bert,gpt2_radford2019language,debertav3_he2021debertav3,roberta_liu2019roberta,gpt3_brown2020language} has consequently become a mainstream paradigm to reduce the training time required for individual tasks, yet with superior performance compared with other methods~\citep{prompt_tuning_lester2021power,prefix_tuning_li2021prefix,adpter_houlsby2019parameter}.

\begin{figure*}[!htb]
    \centering
    \includegraphics[width=.825\linewidth]{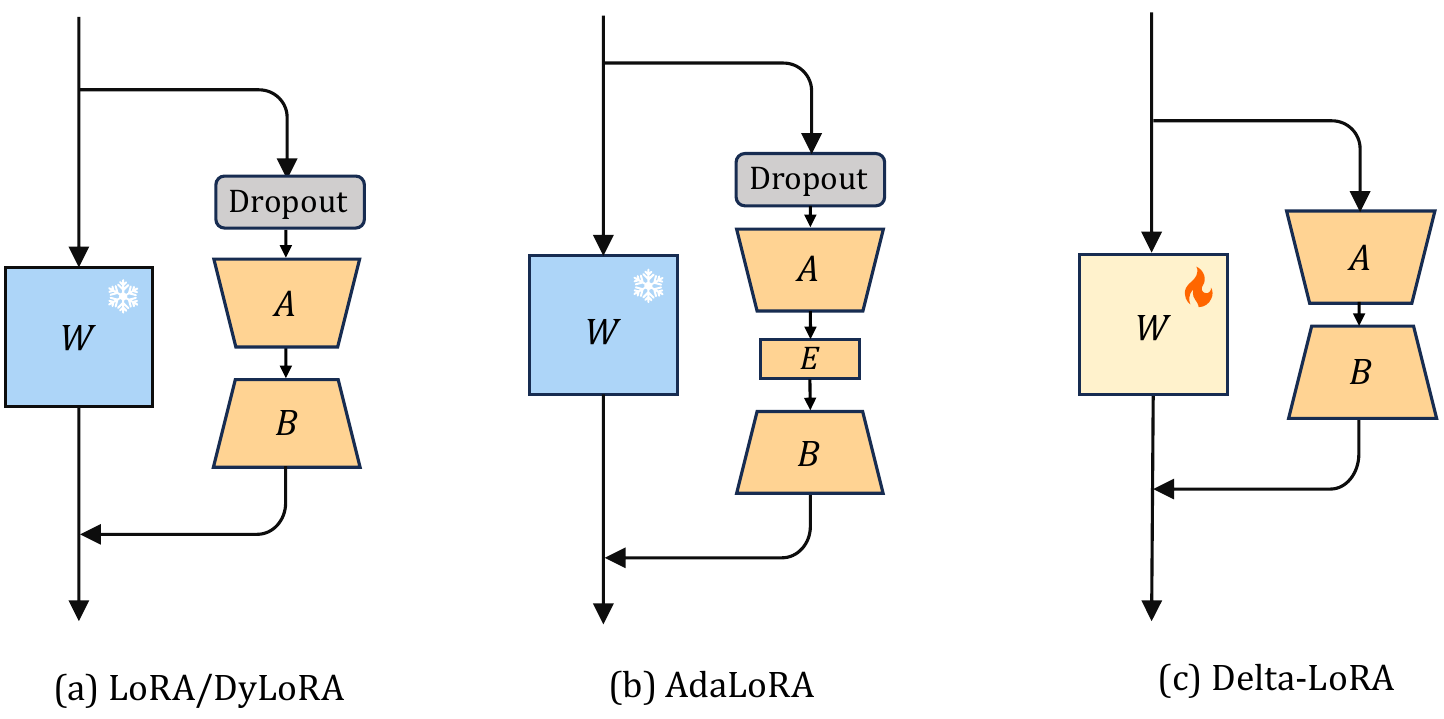}
    \caption{An overview of the proposed \textbf{Delta-LoRA} structure, compared to \textbf{LoRA}, \textbf{DyLoRA} and \textbf{AdaLoRA}. Note that \textbf{DyLoRA} and \textbf{LoRA} basically share the same architecture.  $\bW$ is the pre-trained weight which is frozen (signified by \textcolor[RGB]{172,212,246}{\textbf{blue}}) when performing efficient-parameter fine-tuning in (a) and (b). \textcolor[RGB]{255,182,82}{\textbf{Orange}} trapezoids $\bA$, $\bB$ and $\bE$ denote the trainable parameters. In our proposed Delta-LoRA, the \textcolor[RGB]{252,230,190}{\textbf{light orange}} rectangle means that pre-trained weights can be updated via the delta. \emph{Note that our proposed Delta-LoRA removes the Dropout layer to ensure reasonable delta for pre-trained matrix. }
    }
    \label{fig:lora}
\end{figure*}
However, fine-tuning a LLM with all the learnable parameters (Full Fine-tuning) requires multiple GPUs with high memory demand~\citep{qlora_dettmers2023qlora, lora_hu2022lora}, 
which is unattainable for many companies and research institutions. 
Full fine-tuning poses exceptional challenges to researchers:
with massive parameter size, LLMs already demand more storage space than regular models; Further training exaggerates the GPU memory requirement because common optimizers such as AdamW~\cite{adamw_loshchilov2017decoupled} often maintain several copies of the model parameters, which is 2-3 times of memory overhead.
To this end, a series of methods have been proposed~\citep{dylora_valipour2023dylora,adalora_zhang2022adaptive,prefix_tuning_li2021prefix,IA_3_liu2022few,update_model_on_the_flow_lv2023parameter,qlora_dettmers2023qlora,p_tuning_liu-etal-2022-p,bitfit_zaken2021bitfit,adapterfusion_pfeiffer2021adapterfusion,diff_pruning_guo2021parameter,adpter_houlsby2019parameter,adamix_wang2022adamix}  to reduce memory overhead at the training stage. Some even accelerate the fine-tuning process with only less than $1\%$ trainable parameters. Among these methods, LoRA~\citep{lora_hu2022lora} is the most attractive for its stable performance on broad downstream tasks~\citep{deltatuning_ding2022delta}, no observed overfitting, as well as no extra memory and computation cost at inference.

While LoRA and its successors~\citep{adalora_zhang2022adaptive,dylora_valipour2023dylora} 
have indeed exhibited superior performance in comparison to alternative approaches within the realm of Parameter Efficient Fine-Tuning (PEFT), a substantial \textbf{performance gap} persists when compared to the full fine-tuning, as highlighted in most scenarios~\citep{deltatuning_ding2022delta}. This discrepancy is attributed to the inherent limitation of updating only a fraction of the model's parameters, rendering it inadequate to fit the intricacies presented in the training data. 

To bridge this gap, a reasonable strategy is to introduce more parameters into the optimization process. In this paper, we introduce Delta-LoRA as shown in Fig.~\ref{fig:lora}, a novel PEFT approach that simultaneously updates the pre-trained matrix and two low-rank matrices while maintaining the same memory consumption as the original LoRA. 
Specifically, the pre-trained matrix $\bW$ is updated with the delta of the product of two low-rank matrices in two consecutive iterations ($\triangle \bA\bB=\bA^{(t+1)}\bB^{(t+1)} - \bA^{(t)}\bB^{(t)}$), while two low-rank matrices are updated by the AdamW optimizer automatically. This is based on the mathematical property that $\frac{\partial \gL}{\partial \bW} = \frac{\partial \gL}{\partial \bA\bB}$ and $\triangle \bA\bB$ is a surrogate to direct the update of $\bW$ (see Sec. \ref{sec:Methodology} for details). Since we neither store the gradient of $\bW$ nor use the optimizer to update the pre-trained matrix, the proposed method thus does not yield any extra memory overhead. 
This strategic integration effectively mitigates the sub-optimal representation learning stemming from only updating the two low-rank matrices. Moreover, our approach aligns the update direction of the pre-trained weights with that of the incremental update matrix. 
Furthermore, we discard the Dropout layer in low-rank branches to obtain a more reasonable delta for $\bW$, in order to ensure $\frac{\partial \gL}{\partial \bW} = \frac{\partial \gL}{\partial \bA\bB}$.
The advantages of our proposed method are conspicuous: including the pre-trained weights in the optimization process engenders a broader integration of parameters, thereby enhancing the potential for learning intricate representations.

The main contributions of this paper can be summarized as:
\begin{itemize}[leftmargin=*]
    \item We introduce Delta-LoRA, a novel PEFT method that simultaneously updates the full weight matrix and two low-rank matrices. Delta-LoRA leverages the delta of the product of $\bA$ and $\bB$ to update the pre-trained weights and thus prevent storing the first and the second-order momentums in the optimizer.
    
    \item We analyze the gradient flow of Delta-LoRA and show that the Dropout layer in the low-rank branch makes $\frac{\partial \gL}{\partial \bW} \neq \frac{\partial \gL}{\partial \bA\bB}$. Thus, we remove the Dropout layer in our proposed Delta-LoRA to get reasonable delta for $\bW$. 
    
    \item We conduct comprehensive experiments to show that Delta-LoRA has consistent gains on a broad range of NLP tasks. Additionally, we provide thorough explanations to analyze its superiority and the value contributed by each component.
\end{itemize}

\section{Preliminaries}

\textbf{Transformer-based Models.}
Transformer~\citep{attention_is_all_you_need_vaswani2017attention} adopts the self-attention mechanism instead of recurrence and convolutions, achieving new state-of-the-art in machine translation. 
\citet{vit_dosovitskiy2020image} later proposed the Vision-Transformer (ViT)  architecture which exhibits versatility across various computer vision tasks. Nowadays, the Transformer-based models have become the most popular choice in 
both NLP and Computer Vision~\citep{ALBEF_li2021align,od_carion2020end,segmformer_zheng2021rethinking}.
Transformer typically consists of $L$ stacked blocks, each containing a multi-head attention (MHA) module and a feed-forward network (FFN) module. 
For an input sequence $\bX \in \mathbb{R}^{n\times d}$, the MHA module yields the output $\textup{MHA}(\bX)$, given by:
\begin{equation}
\begin{split}
    \textup{head}_i = \textup{softmax}(\frac{\bX\bW_{Q_i} (\bX\bW_{K_i})^{\top}}{\sqrt{d_k}})\bX\bW_{V_i} \\
    \textup{MHA}(\bX) = \textup{concat}(\textup{head}_1,...,\textup{head}_k)\bW_o,
\end{split}
\end{equation}
where $d_k$ is the scaling factor and set to $d_k = d/k$. $\bW_{K_i}$ $\bW_{Q_i}$, $\bW_{V_i}$ and $\bW_o$ are weight matrices for computation of key, query, value and the output of $\textup{MHA}$, respectively. Besides the MHA module, the FFN is also vital in the Transformer-based model. It stacks two fully connected (FC) layers with an activation function in between. FFN is defined as:
\begin{equation}
    \begin{split}
        \textup{FFN}(\bx) = \bW_{f_2} \textup{ReLU}(\bW_{f_1} \bx + \bb_1) + \bb_2,
    \end{split}
\end{equation}
where $\bx \in \mathbb{R}^{d}$, $\bW_{f_1}$ and $\bW_{f_2}$ are two fully connected layers in FFN, $\bb_1$ and $\bb_2$ are bias terms.

\textbf{Low Rank Adaptation}. Given a pre-trained matrix $\bW \in \mathbb{R}^{c \times d}$, LoRA~\citep{lora_hu2022lora} learns an incremental update $\bigtriangleup \bW$ and decomposes $\bigtriangleup \bW$ into  a matrix multiplication between two low-rank matrices $\bA$ and $\bB$, where $\bA\in \mathbb{R}^{c\times r}$ and $\bB\in \mathbb{R}^{r\times d}$, and $\bigtriangleup \bW = \bA \bB$. Here, the rank $r\ll min(d,c)$. For an input $\bx$ and hidden state $\bh$, LoRA has the following forward process:
\begin{equation}
    \bh = \bW^{*} \bx = \bW \bx + \bigtriangleup \bW \bx = \bW \bx + \frac{\alpha}{r} \bA \bB \bx
\end{equation}
At the beginning of the training stage, $\bA$ is randomly initialized via Kaiming initialization~\citep{kaimingintialization_he2015delving} and $\bB$ is initialized to zero matrix to make sure that the incremental update $\bA\bB=\bf{0}$ at initialization. Besides, LoRA uses hyper-parameters $\alpha$ and $r$ to scale $\bA\bB\bx$.

\section{Related Works}
With the ever-growing parameter scale in current Transformer-based models, fine-tuning such a large language model (LLM) requires considerable number of GPUs equipped with high memory capacity. This is mainly due to the fact that common optimizers such as AdamW~\citep{adamw_loshchilov2017decoupled} requires maintaining three times of extra parameter size (gradients, first-order and second-order momentums).
To bridge this gap, a series of Parameter-Efficient Fine-Tuning (PEFT) methods have been proposed~\citep{lora_hu2022lora,p_tuning_liu-etal-2022-p, autoprompt_shin2020autoprompt, adpter_houlsby2019parameter}.
The Adapter~\citep{adpter_houlsby2019parameter} introduces lightweight trainable parameters between pre-trained layers while keeping the pre-trained weights fixed. Prompt-Tuning~\citep{prompt_tuning_lester2021power} aims to optimize the prompt to achieve comparable performance with fine-tuning for specific task, while Prefix-Tuning optimizes for trainable prefixes and prepends these trainable parameters to each hidden state~\citep{prefix_tuning_li2021prefix}. Despite the notable performance achievements, these methods inevitably introduce extra overhead at the inference stage. 

\citet{lora_hu2022lora} proposed LoRA to utilize the multiplication of two low-rank matrices to model the incremental update of a full-rank matrix. LoRA merges the incremental updates to pre-trained weights after training, thereby avoiding any extra computation overhead during inference. Furthermore, it stands out as one of the most effective PEFT techniques according to \citet{deltatuning_ding2022delta}'s evaluation.
Subsequent to its inception, a series of enhanced methods building upon LoRA was proposed. Notably, G-LoRA~\citep{glora_chavan2023one} leverages a generalized prompt module to fine-tune pre-trained weights resulting in better representations for computer vision tasks. DyLoRA~\citep{dylora_valipour2023dylora} aims to adjust the rank of two lightweight matrices after the training stage. Differing from the conventional approach of maintaining a static rank during training, DyLoRA introduces rank variations to its blocks.
AdaLoRA~\citep{adalora_zhang2022adaptive} emphasizes the disparate importance attributed to distinct weight parameters. This technique intelligently allocates the parameter budget across weight matrices based on their respective importance scores.
Additionally, Q-LoRA~\citep{qlora_dettmers2023qlora} was proposed to further reduce the average memory footprint by quantizing the pre-trained model with 4-bit NormalFloat.
This quantization approach not only preserves the model's efficacy but also effectively alleviates the resource-intensive nature of LLM training and addresses a pertinent concern.

\begin{figure*}[!ht]
    \centering
    \includegraphics[width=.835\linewidth]{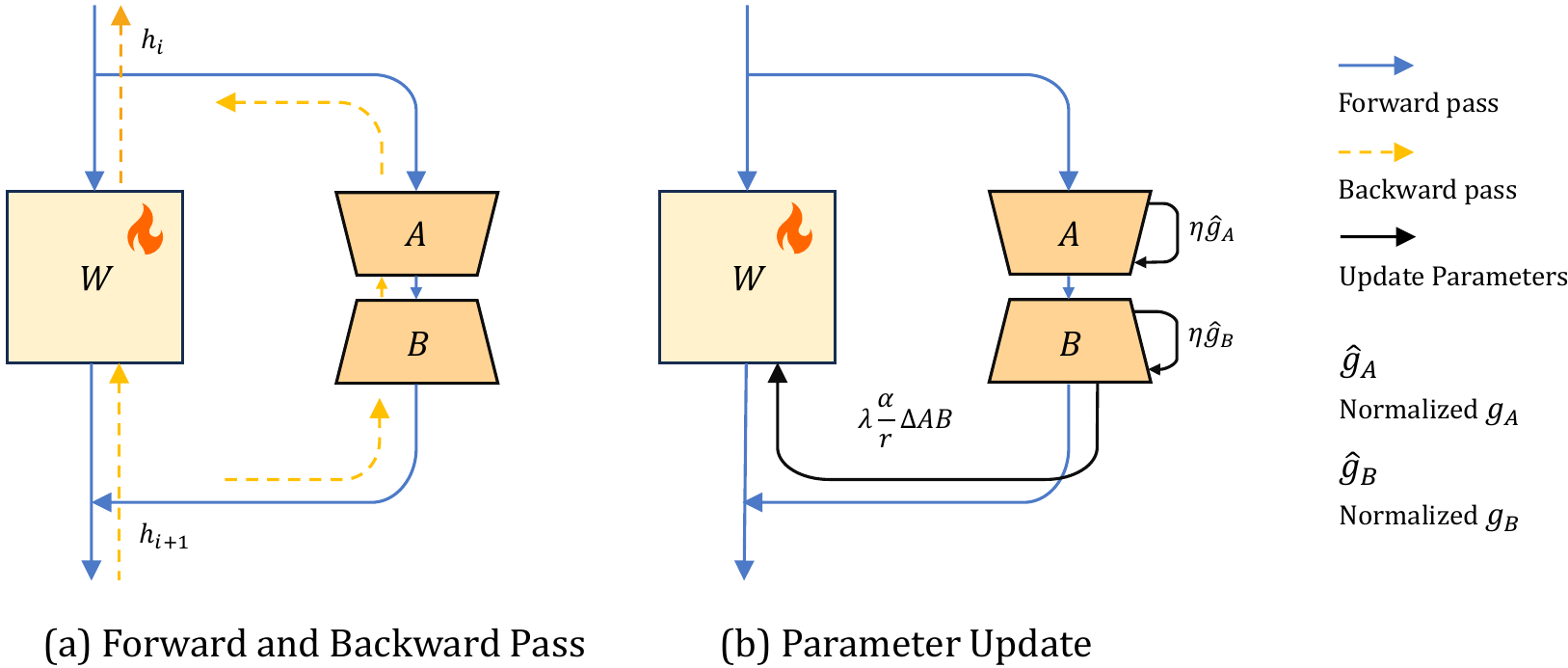}
    \caption{The framework of our proposed Delta-LoRA. The blue arrows represent forward pass while yellow dashed arrows denote backward propagation. The black solid arrows in (b) represent the process of updating the low-rank adaptation matrices $\bA$ and $\bB$ with normalized gradients $\widehat{\bg}_{\bA}$ and $\widehat{\bg}_{\bB}$ multiplied by the learning rate $\eta$, as well as updating the pre-trained weights $\bW$ with the delta matrix $\triangle \bA\bB$ multiplied by the update ratio $\lambda$.}
    \label{fig:lora2}
\end{figure*}

\section{Methodology}\label{sec:Methodology}
This section introduces the novel fine-tuning approach termed as Delta-LoRA. Delta-LoRA encompasses two pivotal designs as shown in Figure~\ref{fig:lora} and Figure \ref{fig:lora2}: (i) It simultaneously updates the full weight matrix ($\bW$) alongside the two low-rank adaptation matrices ($\bA$ and $\bB$), utilizing the delta $(\bA^{(t+1)}\bB^{(t+1)} - \bA^{(t)}\bB^{(t)})$ resulting from incremental updates to refine the pre-trained weights ($\bW$); 
(ii) The Dropout layer as originally integrated within the conventional LoRA module, is excluded in Delta-LoRA. This omission stems from the realization that its presence violates the required assumption$\frac{\partial \gL}{\partial \bW} = \frac{\partial \gL}{\partial \bA\bB}$. 

\subsection{Update the Delta of Low-rank Matrices on Pre-trained Weights}
For an input $\bx$ and its corresponding hidden state $\bh$, LoRA optimizes two low-rank matrices $\bA$ and $\bB$ to learn an incremental update $\bA\bB$ for the pre-trained and fixed weight matrix $\bW$.
Different from previous methods, we argue that $\bW$ also needs to be updated. In this way, we can introduce more learnable parameters to the optimization process for higher learning capability. However, acquiring the normalized gradients (i.e. the gradients after normalization in optimizer) to fine-tune the weight matrix $\bW$ is non-trivial, since the optimizer such as AdamW must maintain at least three extra copies of the parameters (i.e. gradients as well as the first-order and the second-order moments of gradients) in GPU memory.
Intriguingly, we note that the gradients of the loss $\gL$ with respect to matrices $\bA\bB$ and $\bW$ are  precisely identical, under the presumption that the LoRA module exclusively retains matrices $\bA$ and $\bB$, while disregarding the Dropout layer. This correspondence can be formally represented as:
\begin{equation}
\begin{split}
    & \bg_{\bW+\bA\bB} = \frac{\partial \gL}{\partial \bh_{i+1}} \cdot (\frac{\partial \bh_{i+1}}{\partial (\bW + \bA\bB)}) ^ {\top} = \frac{\partial \gL}{\partial \bh_{i+1}} \cdot \bh_i ^ {\top}, \\
    & \bg_{\bW} = \frac{\partial \gL}{\partial \bh_{i+1}} \cdot \frac{\partial \bh_{i+1}}{\partial \bW} ^ {\top} = \frac{\partial \gL}{\partial \bh_{i+1}} \cdot \bh_i ^ {\top}, \\
    & \bg_{\bA\bB} = \frac{\partial \gL}{\partial \bh_{i+1}} \cdot \frac{\partial \bh_{i+1}}{\partial \bA\bB} ^ {\top} = \frac{\partial \gL}{\partial \bh_{i+1}} \cdot \bh_i ^ {\top},\\
    & \Longrightarrow  \bm{\textcolor{blue}{ \bg_{\bW} = \bg_{\bA\bB}}} ,
\end{split}
\label{graident}
\end{equation}\label{eq:gW_gAB}
where $\bh_{i+1} = \bW \bh_i + \bA\bB \bh_i$, $\bh_i$ and $\bh_{i+1}$ are the outputs of the $i$-th layer and the $i$+1-th layer respectively. $\bA\bB$ is the matrix product of the adaptation matrices $\bA$ and $\bB$, $ \gL $ is the loss function, while $\bg_{\bW+\bA\bB}$, $\bg_{\bW}$ and $\bg_{\bA\bB}$ denote the gradients of $\frac{\partial \gL}{\partial {(\bW+\bA\bB)} }$, $\frac{\partial \gL}{\partial \bW}$, and $\frac{\partial \gL}{\partial \bA\bB}$ respectively. 
\large
\begin{algorithm*}
 \caption{Delta-LoRA}
 \label{algorithm_c-lora}
 \begin{algorithmic}
     \item \textbf{Input:} Learning rate $\eta$; weight decay $\beta$; total training iterations $T$; low rank $r$; scale factor $\alpha$; start steps $K$; update ratio $\lambda$.\\
     $\bA$ is initialized by Kaiming Initialization, $\bB=\bf{0}$ and $\bW$ is initialized with pre-trained weights. \\
     \textbf{for} $t=0,...,T-1$ \textbf{do} \\
     \ \ \ \ Sample a mini-batch and compute gradients for \{$\bA$,$\bB$\} in each Delta-LoRA module.\\
     \ \ \ \ Update the first and second moments maintained by the optimizer with the computed gradients, and get the normalized gradients $\widehat{g}_{\bA}$ and $\widehat{g}_{\bB}$.\\
     \ \ \ \ $\bA^{(t+1)} \leftarrow \bA^{(t)}-\eta \widehat{g}_{\bA} - \eta \beta \bA^{(t)}$ \\
     \ \ \ \ $\bB^{(t+1)} \leftarrow \bB^{(t)}-\eta \widehat{g}_{\bB} - \eta \beta \bB^{(t)}$ \\
     \ \ \ \ \textbf{if} $t>K$ \textbf{do} \\
     \ \ \ \ \ \ \ $\bW^{(t+1)} \leftarrow \bW^{(t)} + \lambda \cdot \frac{\alpha}{r} \cdot (\bA^{(t+1)}\bB^{(t+1)} - \bA^{(t)}\bB^{(t)})$ \\
     \ \ \ \ \textbf{end if} \\
     \textbf{end for} \\
     \textbf{Output:} the fine-tuned parameters \{$\bW^{(T)}, \bA^{(T)}, \bB^{(T)}$\}
 \end{algorithmic}

\end{algorithm*}

\normalsize
Equation~\ref{graident} inspires us to use $\bg_{\bA\bB}$ to assimilate $\bg_{\bW}$ when learning the parameter updates for weight matrix $\bW$.
Unfortunately, we are only able to obtain the gradients $\bg_{\bA}$ and $\bg_{\bB}$ rather than $\bg_{\bW}$ during the back-propagation process. Furthermore, the computation of the gradients for $\bA\bB$ is as expensive as for the matrix $\bW$, since both matrices share the same dimensions of $d\times k$, consequently entailing an equivalent GPU memory overhead. 

Considering a typical optimization process, the model updates its parameters by applying the gradient descent: $\bW^{(t+1)} = \bW^{(t)} - \eta {\bg}_{\bW}$, with the parameter update denoted as $\triangle \bW = -\eta {\bg}_{\bW}$, using the learning rate $\eta$. 
Similarly, we regard $-\triangle \bA \bB$ as the gradients for $\bA\bB$ and utilize this matrix as a substitute for $\bg_{\bW}$ according to Equation ~\ref{graident}. Here, we can compute $\triangle \bA \bB$ as:
\begin{equation}
        \triangle \bA\bB = \bA^{(t+1)}\bB^{(t+1)} - \bA^{(t)}\bB^{(t)} 
        = \eta \bA^{(t)} \bg_{\bB} + \eta \bg_{\bA}\bB^{(t)} - \eta^{2} \bg_{\bA}\bg_{\bB}, \\
\end{equation}
where $\bA^{(t)}$, $\bB^{(t)}$ and $\bW^{(t)}$ are the weights of $\bA$, $\bB$ and $\bW$ at the $t$-th step respectively, $\bA^{(t+1)} = \bA^{(t)} - \eta \bg_{\bA}$, $\bB^{(t+1)} = \bB^{(t)} - \eta \bg_{\bB}$ and $\eta$ is the learning rate.
To be precise, $-\triangle \bA \bB$ does not equate directly to $\bg_{\bA \bB}$ and $\bg_{\bW}$ as elaborated in Appendix \ref{expansion}. Nonetheless, $\triangle \bA \bB$ has the capability to symbolize the genuine directions of update for the matrix $\bA\bB$. Based on this assumption, it is reasonable to employ $-\triangle\bA\bB$ as the gradient for directing the update of $\bW$.

Therefore, during the training phase we introduce the matrix $\triangle \bA \bB$ to update the pre-trained weights $\bW$ in the following manner:
\begin{equation}
\label{c_lora}
    \bW^{(t+1)} = \bW^{(t)} + \lambda \cdot \frac{\alpha}{r} \cdot \triangle \bA\bB, \text{where} \ \ \triangle \bA\bB = \bA^{(t+1)}\bB^{(t+1)} - \bA^{(t)}\bB^{(t)}, 
\end{equation}
where $\lambda$ represents the hyper-parameter to trade off the update ratio of $\bA\bB$ and the pre-trained weights $\bW$. The parameter updates for $\bW$ commence after $K$ training iterations. The procedural details of the algorithm are illustrated in Algorithm ~\ref{algorithm_c-lora}.

\textbf{Discussion.} The Delta-LoRA has some important modifications compared to LoRA. Here, we discuss and compare the difference:

\begin{tikzpicture}
    \coordinate (A) at (0,0);
    \coordinate (B) at (14,0);
    \coordinate (C) at (14,2.6);
    \coordinate (D) at (0,2.6);
    
    \fill[gray!10] (A) -- (B) -- (C) -- (D) -- cycle;
    
    \draw[dashed] (7,0) -- (7,2.6);
    
    \node at (3.5,2.25) {\textbf{LoRA}};
    
    \node at (3,1.55) {(1) $\bA^{(t+1)} \leftarrow \frac{\partial\gL(\bx;\bW,\bA^{(t)},\bB^{(t)})}{\partial \bA^{(t)}}$};
    \node at (3,0.65) {(2) $\bB^{(t+1)} \leftarrow \frac{\partial\gL(\bx;\bW,\bA^{(t)},\bB^{(t)})}{\partial \bB^{(t)}}$};

    \node at (10.5,2.25) {\textbf{Delta-LoRA}};
    \node at (10,1.65) {(1) $\bA^{(t+1)} \leftarrow \frac{\partial\gL(\bx;\bW^{(t)},\bA^{(t)},\bB^{(t)})}{\partial \bA^{(t)}}$};
    \node at (10,0.95) {(2) $\bB^{(t+1)} \leftarrow \frac{\partial\gL(\bx;\bW^{(t)},\bA^{(t)},\bB^{(t)})}{\partial \bB^{(t)}}$};
    \node at (10.52,0.30) {(3) $\bW^{(t+1)} \leftarrow \bA^{(t+1)}\bB^{(t+1)} - \bA^{(t)}\bB^{(t)}$};
\end{tikzpicture}
It is obvious that LoRA only updates $\bA$ and $\bB$, and keep $\bW$ frozen, while Delta-LoRA updates $\bA$ and $\bB$ by the optimizer and $\bW$ with the delta of the product of $\bA$ and $\bB$.

\subsection{The structure of our Delta-LoRA}
Both LoRA and its successor AdaLoRA put a Dropout layer before two low-rank matrices $\bA$ and $\bB$. However, this arrangement results in a disparity between the gradient matrices $\bg_{\bW}$ and $\bg_{\bA\bB}$ (or the matrix $\bg_{\bA\bE\bB}$ in the context of AdaLoRA).  The derivation of this disparity can be shown as:
\begin{equation}
\begin{split}
    &g_{\bW} = \frac{\partial \gL}{\partial \bh_{i+1}} \cdot (\frac{\partial \bh_{i+1}}{\partial \bW})^{\top} = \frac{\partial \gL}{\partial \bh_{i+1}} \cdot \bh_i^{\top} \\
    &\neq \bg_{\bA\bB} = \frac{\partial \gL}{\partial \bh_{i+1}} \cdot (\frac{\partial \bh_{i+1}}{\partial \bA\bB})^{\top} = \frac{\partial \gL}{\partial \bh_{i+1}} \cdot \textup{Drop}(\bh_i)^{\top},
\end{split}
\end{equation}
where $\textup{Drop}(\cdot)$ denotes the Dropout layer which leads to $\bg_{\bW} \neq \bg_{\bA\bB}$.
A reasonable choice is to remove the Dropout layer in the low-rank module and activate the Dropout layer between pre-trained layers if overfitting problem occurs.
This modification also brings additional benefits: (1) it can alleviate under-fitting to some extent, thereby enhancing the learned representations of the networks. The rationale behind this improvement lies in the fact that LoRA and its successors formulate low-rank updates for pre-trained weights, involving less than $1\%$ of the complete parameters. However, relying solely on such a small fraction of parameters may not bestow an adequate representation capacity in most cases; (2) This alteration also yields memory-saving benefits. By negating the requirement to store intermediate features, the model curtails the memory consumption. Consequently, there is a reduction in activation memory employed during the back-propagation process.

\section{Experiments}
We evaluate our proposed model fine-tuning method Delta-LoRA with RoBERTa~\citep{roberta_liu2019roberta}, GPT-2~\citep{gpt2_radford2019language} and BART~\citep{bart_lewis2019bart} on a broad set of datasets. Specifically, we train (1) RoBERTa on GLUE benchmark which consists of 8 NLP understanding tasks; (2) GPT-2 on E2E Challenge and WebNLG Challenge 2017 following the setting of \citet{lora_hu2022lora}; and (3) BART on XSum dataset by using the setting provided by \citet{adalora_zhang2022adaptive}. See Appendix \ref{parameter} for more training details on the datasets. We use \emph{PyTorch} to implement our experiments and download the pre-trained weights as well as configuration files from \emph{HuggingFace} \citet{huggingface_wolf2019huggingface}. 

\subsection{Baselines}
We compare our proposed method Delta-LoRA with Fine-Tuning and prior works of LoRA, AdaLoRA, and DyLoRA. For PEFT methods, we only train the incremental updates for $\bW_{\bV}$ and $\bW_{\bQ}$, following the setup as used in LoRA's paper. For Fine-Tuning methods, we use two extra training paradigms: (1) freeze the embedding and train all the other parameters as Fine-Tuning $\dag$; (2) train $\bW_{\bV}$ and $\bW_{\bQ}$ only as Fine-Tuning$\ddag$.

\textbf{Fine-Tuning.} 
In the past few years, fine-tuning has become the mainstream paradigm for both NLP and CV tasks. However, fine-tuning full parameters is subject to potential drawbacks including overfitting and training instability~\citep{huang-etal-2022-sparse}. Therefore, freezing a subset of network layers and fine-tuning the rest has become a popular choice \citep{tan2018survey}. In our experiments, we compare with full fine-tuning, fine-tuning with embedding layers frozen (Fine-tuning $\dag$) and fine-tuning query and value matrices only (Fine-tuning $\ddag$). 

\textbf{LoRA} \citep{lora_hu2022lora} uses multiplication of two low-rank matrices to learn the incremental updates with reduced GPU memory cost. We follow their setups to reproduce experimental results for fair comparison. 

\textbf{DyLoRA} \citep{dylora_valipour2023dylora} randomly chooses a rank $r$ for LoRA modules during learning. 

\textbf{AdaLoRA} \citep{adalora_zhang2022adaptive} focuses on the challenge of determining the optimal rank for incremental updates. It employs an adaptive approach to singular value pruning, tailoring the rank selection to the magnitude of each singular value. Consequently, distinct ranks are employed for different layers.

\begin{table*}[h]\small
\begin{center}
\caption{The evaluation results of our proposed Delta-LoRA and other existing methods on E2E NLG Challenge dataset. $\dag$ indicates fine-tuning all layers except embedding layer. $\ddag$ indicates only fine-tuning weights for query and value. $\P$ means we choose different settings with AdaLoRA: we only tune $\bW_{\bQ}$ and $\bW_{\bV}$ instead of all layers. The best results of Fine-Tuning methods are \underline{underlined}. The best results of PEFT methods are \textbf{boldfaced}.}
\label{tab_eval_e2e}
\setlength{\tabcolsep}{0.95mm}{
\begin{tabular}{c|c|c|ccccc}
\toprule \multirow{2}{*}{Method} &Trainable& Extra Updatable & \multirow{2}{*}{BLEU} & \multirow{2}{*}{NIST} & \multirow{2}{*}{METEOR} & \multirow{2}{*}{ROUGE-L} & \multirow{2}{*}{CIDEr} \\
& Parameters & Parameters & & & & &\\
\midrule
Full Fine-Tuning &354.92M& \XSolidBrush &69.58 & 8.75 & \underline{46.34} & 71.66 & 2.47 \\
Fine-Tuning$\dag$ &305.84M& \XSolidBrush &69.37 & 8.76 & 46.05 & \underline{71.97} & 2.44 \\
Fine-Tuning$\ddag$&48M& \XSolidBrush &\underline{69.77} & \underline{8.84} & 46.29 & 71.96 & \underline{2.49} \\
\bottomrule
LoRA (repr.) &0.375M& \XSolidBrush &69.60 & 8.78 & 45.61 & 71.12 & 2.45  \\
LoRA &\textcolor[RGB]{199,199,199}{0.35M}& \textcolor[RGB]{199,199,199}{\XSolidBrush} & \textcolor[RGB]{199,199,199}{70.4} & \textcolor[RGB]{199,199,199}{8.85} & \textcolor[RGB]{199,199,199}{46.8} & \textcolor[RGB]{199,199,199}{71.8} & \textcolor[RGB]{199,199,199}{2.53}\\
DyLoRA &0.375M& \XSolidBrush & 67.89& 8.50& 44.07& 70.52& 2.26\\
AdaLoRA$\P$ &0.375M& \XSolidBrush & 68.16 & 8.58 & 44.10 & 70.66 & 2.35\\
Delta-LoRA (Ours) &0.375M& \Checkmark\ 48M &\textbf{70.84} & \textbf{8.91} & \textbf{46.47}& \textbf{72.24}& \textbf{2.53}\\
\bottomrule
\end{tabular}
}
\end{center}
\end{table*}

\begin{table*}[h]\small
\begin{center}
\caption{The evaluation results of our proposed Delta-LoRA and other existing methods on WebNLG Challenge 2017 dataset. $\dag$ indicates fine-tuning all layers except embedding layer. $\ddag$ indicates only fine-tuning weights for query and value. $\P$ means we choose different settings with AdaLoRA: we only tune $\bW_{\bQ}$ and $\bW_{\bV}$ instead of all layers. The best results of Fine-Tuning methods are \underline{underlined}. The best results of PEFT methods are \textbf{boldfaced}.}
\label{tab_eval_webnlg}
\setlength{\tabcolsep}{1mm}{
\begin{tabular}{c|c|c|ccc|ccc|ccc}
    \toprule \multirow{2}{*}{Method} &Trainable & 
 Extra Updatable &\multicolumn{3}{c|}{BLEU$\uparrow$} & \multicolumn{3}{c|}{METEOR$\uparrow$}& \multicolumn{3}{c}{TER$\downarrow$} \\
& Parameters & Parameters & S & U & A & S & U &A & S & U & A \\
\midrule
Full Fine-Tuning &354.92M& \XSolidBrush & 61.38 & 45.11 & 54.48 & 0.44 & 0.38 & 0.41 & 0.36 & 0.53 & 0.44 \\
Fine-Tuning$\dag$ &305.84M& \XSolidBrush & 63.53 & 46.66 & 55.92 & 0.45 & 0.39 & 0.42 & 0.34 & 0.49 & 0.41  \\
Fine-Tuning$\ddag$ & 48M & \XSolidBrush & \underline{64.55} & \underline{48.06} & \underline{57.08}  &  \underline{0.46} & \underline{0.39} & \underline{0.43} & \underline{0.33} & \underline{0.47} & \underline{0.40}\\
\bottomrule
LoRA (repr.) &0.375M& \XSolidBrush &62.08 & 46.61 & 55.05 & 0.44 & 0.38 & 0.41 & 0.35 & 0.49 & 0.42  \\
\textcolor[RGB]{199,199,199}{LoRA} &\textcolor[RGB]{199,199,199}{0.375M}& \textcolor[RGB]{199,199,199}{\XSolidBrush} &\textcolor[RGB]{199,199,199}{62.1} & \textcolor[RGB]{199,199,199}{46.7} & \textcolor[RGB]{199,199,199}{55.3} & \textcolor[RGB]{199,199,199}{0.44} & \textcolor[RGB]{199,199,199}{0.38} & \textcolor[RGB]{199,199,199}{0.41} & \textcolor[RGB]{199,199,199}{0.33} & \textcolor[RGB]{199,199,199}{0.46} & \textcolor[RGB]{199,199,199}{0.39}  \\
DyLoRA &0.375M& \XSolidBrush & 58.39 & 46.02 & 52.77 & 0.42 & 0.37 & 0.40 & 0.38 & 0.49 & 0.43 \\
AdaLoRA$\P$ &0.375M& \XSolidBrush & 56.39 & 44.14 & 50.82 & 0.41 & 0.37 & 0.39 & 0.40 & 0.49 & 0.44 \\
Delta-LoRA (Ours) &0.375M&\Checkmark 48M&\textbf{62.87} & \textbf{47.68} & \textbf{55.96} & \textbf{0.45} & \textbf{0.39} & \textbf{0.42} & \textbf{0.34} & \textbf{0.48} & \textbf{0.40}\\
\bottomrule
\end{tabular}
}
\end{center}
\end{table*}

\subsection{Natural Language Generation}
\textbf{Models and Datasets.} We use GPT2-Medium to verify the effectiveness of our Delta-LoRA  on two datasets for data-to-text tasks, including the E2E NLG Challenge~\citep{puzikov-gurevych-2018-e2e} and WebNLG Challenge 2017~\citep{gardent-etal-2017-webnlg}. GPT2-Medium has 354M parameters with 24 Transformer layers. The E2E NLG Challenge dataset contains around 42,000 training examples, 4,600 validation examples, and 4,600 test examples from the restaurant domain. The WebNLG Challenge 2017 contains 21,855 training samples of 9 categories, with a total of 14 categories in the test set. For the text summarization task, we use BART-Large~\citep{bart_lewis2019bart} to verify the effectiveness of our method on XSum dataset~\citep{xsum_narayan2018don}, which consists of 204,045 samples for training, 11,332 samples for validation and 11,332 samples for test. 

\textbf{Implementation Details.} 
In order to compare with LoRA and its successors fairly, we adopt the model setups from LoRA to implement our Delta-LoRA and three PEFT methods. We only learn the low-rank incremental update for $\bW_Q$ and $\bW_V$ in MHA module.
Meanwhile, the training configurations are also selected according to existing baselines in order to make a fair comparison. 
For data-to-text datasets, we use the same training configurations as adopted by LoRA, including the number of training epochs, batch size and etc. We use update ratio $\lambda=2$ and set start steps $K=500$ for Delta-LoRA. More details about Delta-LoRA are listed in the Appendix \ref{parameter}.
For the text-summarization task, we use the implementation of AdaLoRA and adopt the same training configurations. We set the update ratio $\lambda=0.5$ and the start steps $K=1000$ for Delta-LoRA.

\textbf{Experimental Results.}
Table \ref{tab_eval_e2e} shows the results for E2E Challenge dataset on 5 evaluation metrics, demonstrating that our method achieves state-of-the-art performance over 3 baselines and a set of fine-tuning methods. For the BLEU and ROUGE-L metrics, our method obtains 1.24 and 1.13 performance gain compared with LoRA, with 0.13, 0.86 and 0.08 improvement on NIST, METEOR and CIDEr respectively. 
Table \ref{tab_eval_webnlg} demonstrates that Delta-LoRA outperforms baselines on BLEU score for WebNLG Challenge 2017 dataset, with 0.79, 1.08 and 0.91 improvement on Seen, Unseen and All test data, respectively. Additionally, for the METEOR and TER evaluation metrics, Delta-LoRA also achieves state-of-the-art performance, with 0.01 and 0.02 improvement over LoRA on all data.
For the text-summarization task, the test results are shown in Table \ref{tab_eval_xsum}, which demonstrates that our method achieves state-of-the-art results across 3 parameter-efficient methods on 4 evaluation metrics. 

\begin{table}[htp!]\small
\begin{center}
\caption{The evaluation results of our proposed Delta-LoRA and other existing methods on XSum dataset. $\dag$ indicates fine-tuning all layers except the embedding layer. $\ddag$ indicates only fine-tuning weights for query and value. $\P$ means we choose different settings with AdaLoRA: we only tune $\bW_{\bQ}$ and $\bW_{\bV}$ instead of all layers. The best results of Fine-Tuning methods are \underline{underlined}. The best results of PEFT methods are \textbf{boldfaced}.}
\label{tab_eval_xsum}
\setlength{\tabcolsep}{1mm}{
\begin{tabular}{c|c|c|cccc}
\toprule \multirow{2}{*}{Method} &Trainable & Extra Updatable &\multirow{2}{*}{Rouge-1} & \multirow{2}{*}{Rouge-2} & \multirow{2}{*}{Rouge-L} & \multirow{2}{*}{Rouge-Sum} \\
& Parameters & Parameters & & & & \\
\midrule
Full Fine-Tuning &387.5M& \XSolidBrush &  \underline{45.36} & \underline{22.16} & \underline{37.23} & \underline{37.24} \\
Fine-Tuning$\dag$  & 338.4M & \XSolidBrush & 45.04 & 22.05 & 36.92 & 36.94 \\
Fine-Tuning$\ddag$  & 72M & \XSolidBrush & 44.95 & 21.43 & 36.35 & 36.37 \\
\bottomrule
LoRA &0.45M& \XSolidBrush &43.27 & 20.13 & 35.12 & 35.12  \\
DyLoRA &0.56M& \XSolidBrush &41.84 & 18.76 & 33.56 & 33.57  \\
AdaLoRA$\P$ &0.56M& \XSolidBrush & 42.91 & 19.76 & 34.71 & 34.72  \\
Delta-LoRA (Ours)&0.56M & \Checkmark 72M & \textbf{43.49} & \textbf{20.23} & \textbf{35.26} & \textbf{35.26} \\
\bottomrule
\end{tabular}
}
\end{center}
\end{table}

\subsection{Natural Language Understanding}
\textbf{Models and Datasets.} We use RoBERTa-base to evaluate the performance of our proposed method, prior works and two fine-tuning methods. We choose the GLUE benchmark in order to conduct fair and clear comparison with LoRA and its successors. This benchmark consists of 8 datasets~\citep{wang2018glue}, including classification tasks, similarity and paraphrase tasks and natural language inference tasks.

\textbf{Implementation Details.}
We use RoBERTa-base with 118M parameters to conduct our experiments and to compare our method with the baselines. 
We mostly adopt the same training configurations of LoRA except for the input length, which is reduced from 512 to 256 in order to reduce memory cost and to accelerate the training process.
We set the rank to 8 and the target rank to 6 for AdaLoRA and choose the rest of hyper-parameters according to the characteristics of different tasks. For Delta-LoRA, we set the update ratio $\lambda$ to 0.5 and choose different start steps $K$ according to warmup steps used in individual tasks.

\begin{table*}[htp!] \small
\begin{center}
\caption{The evaluation results of our proposed Delta-LoRA and other existing methods on GLUE benchmark. We report the overall (matched and mismatched) accuracy for MNLI, Matthew’s correlation for CoLA, Pearson correlation for STS-B, and accuracy for other tasks. $\dag$ indicates fine-tuning all layers except the embedding layer. $\ddag$ indicates only fine-tuning weights for query and value. $\P$ means we choose different settings with AdaLoRA: we only tune $\bW_{\bQ}$ and $\bW_{\bV}$ instead of all layers. The best results of Fine-Tuning methods are \underline{underlined}. The best results of PEFT methods are \textbf{boldfaced}.}
\label{tab_eval_glue}
\setlength{\tabcolsep}{0.5mm}{
\begin{tabular}{c|c|c|ccccccccc}
\toprule \multirow{2}{*}{Method} &Trainable& Extra Updatable &\multirow{2}{*}{MNLI} & \multirow{2}{*}{SST-2} & \multirow{2}{*}{MRPC} & \multirow{2}{*}{CoLA} & \multirow{2}{*}{QNLI} & \multirow{2}{*}{QQP} & \multirow{2}{*}{RTE} & \multirow{2}{*}{STS-B} & \multirow{2}{*}{AVG} \\
& Parameters & Parameters & & & & & & & & \\
\midrule
Full Fine-Tuning &118.87M& \XSolidBrush &  87.51 &  94.26 &  88.23 &  \underline{64.57} &  92.73 &  \underline{91.96}  &  84.11  &  \underline{90.56} &  86.74 \\
Fine-Tuning$\dag$ &82.05M& \XSolidBrush &  \underline{87.58} &  94.02 &  \underline{89.95} &  62.99 &  92.73 &  91.90  &  \underline{86.64}  &  90.22 &  \underline{87.01} \\
Fine-Tuning$\ddag$ & 13.5M & \XSolidBrush & 87.48 & \underline{95.06} & 89.21 & 61.07 & \underline{92.76} & 91.19 & 84.83 & 89.85 & 86.43
\\
\bottomrule
LoRA &0.28M& \XSolidBrush &  87.40 &  94.62 & 89.97 & 63.17 & 93.02 & 90.67 & 86.64 & 91.54 & 87.12  \\
DyLoRA &0.28M& \XSolidBrush & 86.33 & 94.26 & 89.46 & 61.12 & 92.22 & 90.17 & 84.47 & 91.06 & 86.14  \\
AdaLoRA$\P$ &0.28M& \XSolidBrush & 87.34 & 94.49 & \textbf{90.19} & 61.64& 93.08 & 90.14 & 85.19 & 91.16 & 86.65 \\
Delta-LoRA (Ours) &0.28M & \Checkmark 13.5M & \textbf{87.50} & \textbf{95.06} & \textbf{90.19} & \textbf{63.82} & \textbf{93.09} & \textbf{90.87} & \textbf{87.00} & \textbf{91.57} & \textbf{87.38} \\
\bottomrule
\end{tabular}
}
\end{center}
\end{table*}

\textbf{Experimental Results.}
We compare our method with prior PEFT works. According to Table  \ref{tab_eval_glue}, our method outperforms existing methods on all 8 tasks in GLUE benchmark.
Among these tasks, our method demonstrates significant improvement on SST-2, CoLA and RTE. This is mainly due to the fact that these datasets contain less training data, which hinders the model's capacity to effectively acquire a robust representation when using prior fine-tuning methods.
Delta-LoRA also achieves decent performance on the rest of the datasets, including MNLI, MRPC, QNLI as well STS-B, which proves that our method is stable and reliable across different settings. Interestingly, we find that fine-tuning a small number of pre-trained parameters can bring pronounced performance gain, which proves that our improvements over other PEFT methods are partially due to the fact that we only adjust a small number of pre-trained parameters 
while inheriting the generalization capability of the pre-trained model.

\subsection{Comprehensive Understanding of Delta-LoRA}

\begin{table*}[h]\small
\begin{center}
\caption{The ablation study of our proposed Delta-LoRA on E2E Challenge dataset demonstrates the importance of each component. The best results are \textbf{boldfaced}.}
\label{tab_eval_ablation}
\setlength{\tabcolsep}{0.75mm}{
\begin{tabular}{c|c|c|ccccc}
\toprule \multirow{2}{*}{Method} &Trainable & Extra Updatable &\multirow{2}{*}{BLEU} & \multirow{2}{*}{NIST} & \multirow{2}{*}{METEOR} & \multirow{2}{*}{ROUGE-L} & \multirow{2}{*}{CIDEr} \\
& Parameters & Parameters & & & & &  \\
\midrule
LoRA (repr.) &0.375M& \XSolidBrush & 69.60 & 8.78 & 45.61 & 71.12 & 2.45 \\
Delta-LoRA +  LoRA Module& 0.375M & \Checkmark 48M & 70.29 &8.88 &46.38 & 71.88 & 2.51 \\
Delta-LoRA &0.375M & \Checkmark 48M &\textbf{70.84} & \textbf{8.91} & \textbf{46.47} & \textbf{72.24} & \textbf{2.53} \\
\bottomrule
\end{tabular}
}
\end{center}
\end{table*}

\begin{table*}[h]\small
\begin{center}
\caption{The ablation study of our proposed Delta-LoRA to eliminate the impact of hyper-parameter $\lambda$ on E2E Challenge dataset. The best results are \textbf{boldfaced}.}
\label{tab_eval_large_lr}
\setlength{\tabcolsep}{0.95mm}{
\begin{tabular}{c|c|c|ccccc}
\toprule \multirow{2}{*}{Method} & Learning & \multirow{2}{*}{ \ \ \ \ $\lambda$\ \ \ \ \ } &\multirow{2}{*}{BLEU} & \multirow{2}{*}{NIST} & \multirow{2}{*}{METEOR} & \multirow{2}{*}{ROUGE-L} & \multirow{2}{*}{CIDEr} \\
&Rate & & & &  \\
\bottomrule
LoRA (repr.) & 2e-4 & - & 69.60 & 8.78 & 45.61 & 71.12 & 2.45 \\
LoRA (repr.) & 6e-4 & - & 69.63 & 8.79 & 45.70 & 71.55 & 2.39 \\
Delta-LoRA & 2e-4 & 2 &\textbf{70.84} & \textbf{8.91} & \textbf{46.47} & \textbf{72.24} & \textbf{2.53} \\
\bottomrule
\end{tabular}
}
\end{center}
\end{table*}

\textbf{Ablation study.}
To better understand the contribution of our modified LoRA module (i.e. Delta-LoRA module) and the effectiveness of our update algorithm, we conduct studies on E2E Challenge dataset with GPT2-medium. As shown in Table \ref{tab_eval_ablation}, only updating the pre-trained matrices with delta of low-rank update can indeed achieve performance improvement, while 
further discarding the dropout in Delta-LoRA module obtains the best performance. This observation confirms the indispensable role played by each component within our proposed methodology. 
We have devised an experiment to further differentiate whether the performance enhancement stems from the inherent characteristics of our method rather than solely from the substantial update magnitude.
According to our algorithm, we update the parameters of both pre-trained and low-rank matrices, which can arose the doubt of whether the improvement is caused by updating larger $\triangle\bA\bB$ on the weights instead of introducing more parameters into the optimization process. To answer this question, we design an experiment with results shown in Table \ref{tab_eval_large_lr} to prove the effectiveness of our method. We scale the learning rate of LoRA from 2e-4 to 6e-4 making sure that $\bW + \bA\bB$ can be updated with $3\times \triangle\bA\bB$, which is equivalent to Delta-LoRA when $\lambda$ is set to 2. We find that even by updating with $3 \times \triangle \bA\bB$ on $\bA\bB$, the performance is still not comparable with Delta-LoRA. This experiment further proves that introducing more parameters into the optimization process can force model to learn better representation.

\begin{figure*}[!htb]
    \includegraphics[width=1.15\linewidth]{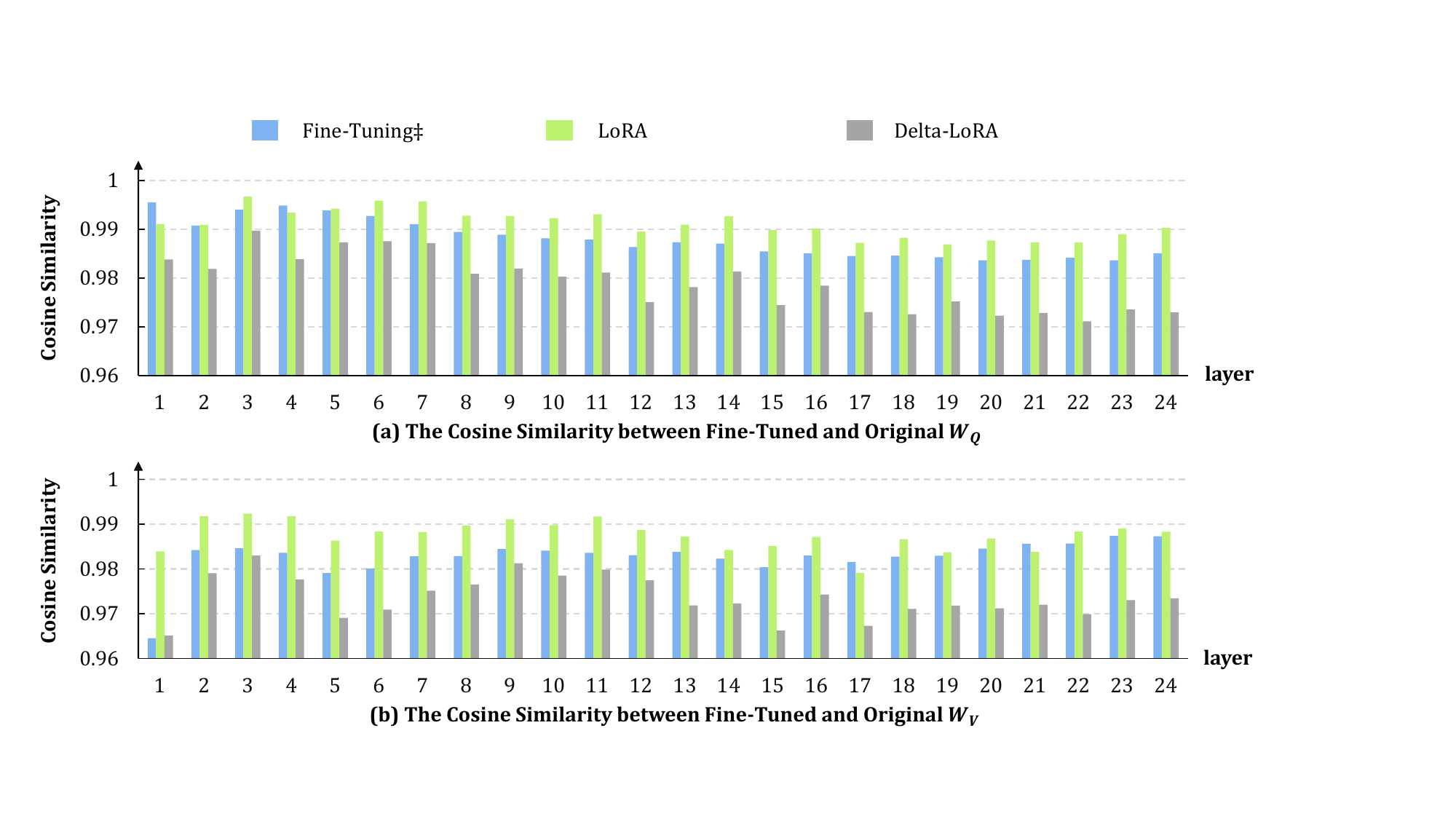}
    \vspace{-0.5in}
    \caption{The comparison of Fine-Tuning$\ddag$, LoRA as well as Delta-LoRA for the cosine similarity between the fine-tuned parameters and the original pre-trained parameters in each transformer block. \emph{Higher value means higher similarity.}}
    \label{fig:similarity}
\end{figure*}

\textbf{The cosine similarity between fine-tuned and the pre-trained parameters to measure learning effects.} 
We conduct a comparative analysis of three methods including Fine-Tuning$\ddag$, LoRA and Delta-LoRA, in order to elucidate the reasons behind Delta-LoRA's superior performance.
We conduct experiments on E2E Challenge dataset, fine-tune or learn incremental updates for the $\bW_{\bQ}$ and $\bW_{\bV}$. We set the learning rate to $2e-4$ and train for 5 epochs.
Subsequently, we use the final checkpoint for conducting comparisons.
As depicted in Figure \ref{fig:similarity}, it is evident that LoRA exhibits the highest similarity across the majority of transformer blocks. This observation suggests that LoRA primarily modifies the matrix $\bW^{*}=\bW+\bA\bB$ within a limited range.
Nevertheless, Delta-LoRA showcases the lowest cosine similarity, underscoring that our approach induces the most significant modifications to the final matrix $\bW^{*}$.
Due to this property, our approach can effectively stimulate the model to acquire better representations, leading to state-of-the-art  performance across all four PEFT methods.
This observation further aligns with the evaluation results in Table \ref{tab_eval_e2e}: Delta-LoRA achieves the best performance among the three methods, whereas LoRA is slightly worse than Fine-Tuning$\ddag$.

\section{Conclusion}
In this paper, we have introduced Delta-LoRA, a novel method to simultaneously update the full weight matrix and two low-rank matrices. Delta-LoRA leverages the delta $(\bA^{(t+1)}\bB^{(t+1)} - \bA^{(t)}\bB^{(t)})$ to update the pre-trained weights ($\bW$).
In this way, we introduce more learnable parameters into the optimization process such that the model can learn a better representation with comparable memory cost as LoRA. Meanwhile, we identify the Dropout layer in the low-rank branch to be unnecessary according to the gradient flow.
We also provide thorough analysis of our
method to understand its effectiveness and robustness. Extensive experiments on a broad range of NLP tasks are conducted to empirically verify the effectiveness of our Delta-LoRA. 

\clearpage
\bibliography{iclr2024_conference}
\bibliographystyle{iclr2024_conference}

\appendix

\clearpage
\section{Appendix}
\label{sec:appendix}

\subsection{The Expansion of $\triangle$ \emph{\textbf{AB}} }
\label{expansion}
In the real training process, we need to consider a variety of training arguments, such as optimizer and the regularization for $\triangle\bA\bB$. Suppose that we use the AdamW \citep{adamw_loshchilov2017decoupled} and $L_{2}$ regularization, the $\triangle \bA \bB$ can be expanded in the following equation:
\begin{equation}
    \begin{split}
        \triangle \bA\bB &= \bA^{(t+1)}\bB^{(t+1)} - \bA^{(t)}\bB^{(t)} \\
        &=(\bA^{(t)} - \eta \widehat{g}_{\bA} - \eta \beta \bA^{(t)}) \cdot (\bB^{(t)} - \eta \widehat{g}_{\bB} - \eta \beta \bB^{(t)}) - \bA^{(t)}\bB^{(t)}\\
        &= \bA^{(t)}\bB^{(t)} -\eta \bA^{(t)}\widehat{g}_{\bB} - \eta \beta \bA^{(t)}\bB^{(t)} -\eta \widehat{g}_{\bA}\bB^{(t)} + \eta^{2}\widehat{g}_{\bA} \widehat{g}_{\bB} + \eta^{2}\beta \widehat{g}_{\bA}\bB^{(t)} \\
        &-\eta\beta\bA^{(t)}\bB^{(t)} + \eta^{2}\beta\bA^{(t)}\widehat{g}_{\bB} +\eta^{2}\beta^{2}\bA^{(t)}\bB^{(t)} - \bA^{(t)}\bB^{(t)} \\
        &= -\eta \bA^{(t)}\widehat{g}_{\bB} - \eta \beta \bA^{(t)}\bB^{(t)} -\eta \widehat{g}_{\bA}\bB^{(t)} + \eta^{2}\widehat{g}_{\bA} \widehat{g}_{\bB} + \eta^{2}\beta \widehat{g}_{\bA}\bB^{(t)} \\
        &-\eta\beta\bA^{(t)}\bB^{(t)} + \eta^{2}\beta\bA^{(t)}\widehat{g}_{\bB} +\eta^{2}\beta^{2}\bA^{(t)}\bB^{(t)} \\
        & \approx  -\eta \bA^{(t)}\widehat{g}_{\bB} - \eta \widehat{g}_{\bA}\bB^{(t)}  
    \end{split}
\end{equation}
where $\eta$ is the learning rate, $\beta$ is weight decay. 
What's more, for pre-trained weight $\bW$, $\triangle \bW=\eta \widehat{g}_{\bW} + \eta \beta \bW^{(t)}$. As a consequence,  $\triangle\bA\bB$ is not equal to $\triangle \bW$ in the training process.

\subsection{The Parameter Sensitivity Study}

\begin{table*}[h]
\begin{center}
\caption{The parameter sensitivity study of 
update ratio $\lambda$ for our proposed Delta-LoRA on E2E Challenge dataset. The best results are \textbf{boldfaced}.}
\label{tab_eval_ablation2}
\setlength{\tabcolsep}{4mm}{
\begin{tabular}{c|ccccc}
\toprule   \ \ \ \ $\lambda$ \ \ \ \ & BLEU & NIST& METEOR& ROUGE-L & CIDEr \\
\bottomrule
0 & 68.94& 8.73 & 45.27 & 70.81 & 2.41 \\
1& 69.77 & 8.81 & 45.99 & 71.58 & 2.46 \\
2& \textbf{70.84} & \textbf{8.91} & \textbf{46.47} & \textbf{72.24} &\textbf{2.53} \\
3& 70.14 & 8.84 & 46.39 & 71.45 & 2.45 \\
4& 70.03 & 8.83 & 46.21 & 71.56 & 2.47 \\
5& 70.13 & 8.85 & 46.35 & 71.72 & 2.48 \\
\bottomrule
\end{tabular}
}
\end{center}
\end{table*}

\textbf{Parameter Sensitivity.}
Here, we explore the hyper-parameter $K$ in Algorithm \ref{algorithm_c-lora} and  $\lambda$ in Equation \ref{c_lora}. For the hyper-parameter $K$, we select it from 0 to 1000 with the interval of 100. From Table \ref{tab_eval_ablation4}, we find that our Delta-LoRA could not bring in any improvement before $K=400$, and it will keep a relatively good performance when $K$ is larger than 500.
What is more, we choose different numbers for $\lambda$, ranging from 0 to 5. According to Table \ref{tab_eval_ablation2}, the 5 metrics rise rapidly after $\lambda=0$ and reach best at $\lambda=2$, while the performance has small drops on 5 evaluation scores if $\lambda$ is chosen from 3 to 5.

\begin{table*}[h]
\begin{center}
\caption{The parameter sensitivity study of start steps $K$ for our proposed Delta-LoRA on E2E Challenge dataset. The best results are \textbf{boldfaced}.}
\label{tab_eval_ablation4}
\setlength{\tabcolsep}{4mm}{
\begin{tabular}{c|ccccc}
\toprule  \ \ \ \ $K$ \ \ \ \  & BLEU & NIST& METEOR& ROUGE-L & CIDEr \\
\bottomrule
0 & 69.10 & 8.75& 45.54 & 71.31 & 2.41 \\
100 & 69.97 & 8.84 & 46.07 & 71.40 & 2.46 \\
200 & 69.72 & 8.83 & 45.82 & 71.41 & 2.43 \\
300 & 69.73 & 8.86 & 45.98 & 71.09 & 2.46 
 \\
400 & 70.18 & 8.89 & 46.30 & 71.66 & 2.49 \\
500& 70.84 & 8.91 & 46.47 & \textbf{72.24} & \textbf{2.53} \\
600 & 70.38 & 8.86 & 46.38 & 71.70 & 2.47 \\
700 & 70.61 & 8.89 & 46.43 & 72.13 & 2.51 \\
800 & 70.70 & 8.89 & 46.30 & 71.97 & 2.51\\
900 & \textbf{71.00} & \textbf{8.92} & \textbf{46.47} & 72.04 & 2.52 \\
1000 & 70.87 & 8.89 & 46.31 & 72.06 & 2.50 \\
\bottomrule
\end{tabular}
}
\end{center}
\end{table*}

\subsection{Hyper-Parameter Used in Our Experiments}
We report the hyper-parameter that used in our experiments. Table \ref{tab_hyper_parameter1} and Table \ref{tab_hyper_parameter2} show the hyper-parameter that  we used for the training and evaluation on E2E Challenge and WebNLG Challenge 2017 dataset. The Table \ref{tab_hyper_parameter3} and Table \ref{tab_hyper_parameter4} are the training and evaluation hyper parameter for XSum dataset, and the Table \ref{tab_training_hyper_parameter_glue} consists of hyper-parameters for 8 datasets in GLUE benchmark. 
\begin{table*}[h]
\begin{center}
\caption{The training hyper-parameter used for E2E Challenge and WebNLG Challenge 2017 dataset.}
\label{tab_hyper_parameter1}
\setlength{\tabcolsep}{1.5mm}{
\begin{tabular}{c|cc}
\toprule  Hyper-Parameter & E2E Challenge & WebNLG Challenge 2017 \\
\bottomrule
Learning Rate $\eta$ & 2e-4 & 2e-4 \\
Batch Size& 8 & 8 \\
Number of Epochs& 5 & 5  \\
Weight Decay $\beta$ & 0.01 & 0.01  \\
Resid\_pdrop & 0 & 0.09 \\
Attn\_pdrop  & 0 & 0.09  \\
Embd\_pdrop  & 0 & 0  \\
Label Smooth & 0 & 0  \\
Start Steps $K$ & 500 & 500 \\
Update Ratio $\lambda$ & 2 & 5 \\
Rank $r$ & 4 & 4 \\
Alpha $\alpha$ & 32 & 32 \\
Trainable Matrices & $\bW_Q$,$\bW_V$ & $\bW_Q$,$\bW_V$ \\
LR Scheduler & Linear & Linear \\
Warmup Steps & 500 & 500 \\
\bottomrule
\end{tabular}
}
\end{center}
\end{table*}

\begin{table*}[h]
\begin{center}
\caption{The hyper-parameter for evaluation used for E2E Challenge and WebNLG Challenge 2017 dataset.}
\label{tab_hyper_parameter2}
\setlength{\tabcolsep}{1.5mm}{
\begin{tabular}{c|cc}
\toprule  Hyper-Parameter & E2E Challenge & WebNLG Challenge 2017 \\
\bottomrule
Beam Size& 10 & 5 \\
Penalty& 0.8 & 1.0  \\
No Repeat Ngram Size& 4 & 4  \\
\bottomrule
\end{tabular}
}
\end{center}
\end{table*}


\begin{table*}[h]
\begin{center}
\caption{The training hyper-parameter used for XSum dataset.}
\label{tab_hyper_parameter3}
\setlength{\tabcolsep}{4mm}{
\begin{tabular}{c|c}
\toprule  Hyper-Parameter & Xsum \\
\bottomrule
Learning Rate $\eta$ & 2e-4  \\
Batch Size& 64 \\
Number of Epochs& 25\\
Weight Decay $\beta$ & 0\\
Activation Dropout & 0 \\
Dropout  & 0\\
Classifier Dropout  & 0\\
Start Steps $K$ & 1000 \\
Update Ratio $\lambda$ & 0.5\\
Rank $r$ & 4\\
Alpha $\alpha$ & 32\\
Trainable Matrices & $\bW_Q$, $\bW_V$ \\
LR Scheduler & Linear \\
Warmup Steps & 3000\\
\bottomrule
\end{tabular}
}
\end{center}
\end{table*}
\begin{table*}[h]
\begin{center}
\caption{The hyper-parameter for evaluation used for XSum dataset.}
\label{tab_hyper_parameter4}
\setlength{\tabcolsep}{4mm}{
\begin{tabular}{c|c}
\toprule  Hyper-Parameter & Xsum \\
\bottomrule
Beam Size& 8 \\
Penalty & 1.0 \\
No Repeat N-gram Size& 4 \\
\bottomrule
\end{tabular}
}
\end{center}
\end{table*}

\begin{table*}[htp!]
\small
\begin{center}
\caption{The training hyper-parameters of our proposed Delta-LoRA on GLUE benchmark. We adopt the most of hyper-parameters in LoRA's paper and implement our method based on the codes given by LoRA's repository.}
\label{tab_training_hyper_parameter_glue}
\setlength{\tabcolsep}{0.95mm}{
\begin{tabular}{c|cccccccc}
\toprule Hyper-Parameter & MNLI & SST-2 & MRPC &CoLA & QNLI & QQP & RTE & STS-B \\
\midrule
Learning Rate $\eta$ &  5e-4 &  5e-4 &  4e-4 &  4e-4 &  4e-4 &  4e-4  &  4e-4  &  4e-4 \\
Batch Size &  128 &  128 &  128 &  64 &  128 & 128  &  128  &  128 
\\
Number of Epochs & 30 & 60 & 30 & 80 & 25 & 25 & 80 & 40 \\
Weight Decay $\beta$ & 0.1 & 0.1 & 0.1 & 0.1 & 0.1 & 0.1 & 0.1 & 0.1 \\
Max Sequence Length &  256 &  256 & 256 & 256 & 256 & 256 & 512 & 256 \\
Start Steps $K$ & 2000 & 400 & 10 & 100 & 800 & 400 & 200 & 200 \\
Update Ratio $\lambda$ & 0.5 & 0.5 & 0.5 &  0.5 & 0.5 & 0.5 & 0.5 & 0.5 \\
Rank $r$ & 8 & 8 & 8 & 8 & 8 & 8 & 8 & 8 \\
Alpha $\alpha$ & 16 & 16 & 16 & 16 & 16 & 16 & 16 & 16 \\
LR Scheduler & Linear & Linear & Linear & Linear & Linear & Linear & Linear & Linear \\
Trainable Matrices & $\bW_Q$,$\bW_V$ & $\bW_Q$,$\bW_V$ & $\bW_Q$,$\bW_V$ & $\bW_Q$,$\bW_V$ & $\bW_Q$,$\bW_V$ & $\bW_Q$,$\bW_V$ & $\bW_Q$,$\bW_V$ & $\bW_Q$,$\bW_V$ \\
Warmup Ratio & 0.06 & 0.06 & 0.06 & 0.06 & 0.06 & 0.06 & 0.06 & 0.06 \\
\multirow{2}{*}{Evaluation Metrics} & \multirow{2}{*}{Accuracy} & Matthews  &\multirow{2}{*}{Accuracy}&Matthews&\multirow{2}{*}{Accuracy}&\multirow{2}{*}{Accuracy}&\multirow{2}{*}{Accuracy}&\multirow{2}{*}{Pearson} \\
& &Correlation &&Correlation&&&& \\
\bottomrule
\end{tabular}
}
\end{center}
\end{table*}


\label{parameter}

\end{document}